\documentclass[conference]{IEEEtran}
\IEEEoverridecommandlockouts
\usepackage{cite}
\usepackage{amsmath,amssymb,amsfonts}
\usepackage{graphicx}
\usepackage{textcomp}
\usepackage{xcolor}
\def\BibTeX{{\rm B\kern-.05em{\sc i\kern-.025em b}\kern-.08em
    T\kern-.1667em\lower.7ex\hbox{E}\kern-.125emX}}

\usepackage{amsmath}
\usepackage{algorithm}
\usepackage{siunitx}
\usepackage[noend]{algpseudocode}
\usepackage{multicol}
\usepackage{wrapfig}
\usepackage{subfig}

\usepackage{listings}

\def\BibTeX{{\rm B\kern-.05em{\sc i\kern-.025em b}\kern-.08em
    T\kern-.1667em\lower.7ex\hbox{E}\kern-.125emX}}
\begin{document}

\title{Root Cause Detection Among Anomalous Time Series Using Temporal State Alignment}

\author{\IEEEauthorblockN{Sayan Chakraborty}
\IEEEauthorblockA{\textit{Zillow Group} \\
Seattle, USA \\
sayanc@zillowgroup.com}
\and
\IEEEauthorblockN{Smit Shah}
\IEEEauthorblockA{\textit{Zillow Group} \\
Seattle, USA \\
smits@zillowgroup.com}
\and
\IEEEauthorblockN{Kiumars Soltani}
\IEEEauthorblockA{\textit{Zillow Group} \\
Seattle, USA \\
kiumarss@zillowgroup.com}
\and
\IEEEauthorblockN{Anna Swigart}
\IEEEauthorblockA{\textit{Zillow Group} \\
Seattle, USA \\
annasw@zillowgroup.com}
}

\maketitle

\begin{abstract}
The recent increase in the scale and complexity of software systems has introduced new challenges to the time series monitoring and anomaly detection process. A major drawback of existing anomaly detection methods is that they lack contextual information to help stakeholders identify the cause of anomalies. This problem, known as \textit{root cause detection}, is particularly challenging to undertake in today's complex distributed software systems since the metrics under consideration generally have multiple internal and external dependencies. Significant manual analysis and strong domain expertise is required to isolate the correct cause of the problem. \par
In this paper, we propose a method that isolates the root cause of an anomaly by analyzing the patterns in time series fluctuations. Our method considers the time series as observations from an underlying process passing through a sequence of discretized hidden states. The idea is to track the propagation of the effect when a given problem causes unaligned but homogeneous shifts of the underlying states. We evaluate our approach by finding the root cause of anomalies in Zillow’s clickstream data by identifying causal patterns among a set of observed fluctuations. 
\end{abstract}


\section{Introduction}
Over the past decade, advancements in cloud computing technologies and distributed data infrastructures have enabled us to tackle critical problems in nearly every field with data-driven approaches. However, it has also caused the software stack to become increasingly complex with many moving pieces distributed among geographically distant data centers. Therefore, it is critical to design scalable monitoring pipelines that can effectively track different metrics and identify inevitable issue with low-latency. \par
A major task in monitoring such complex system is to identify the root cause of a problem. This is particularly challenging since a typical data-driven pipeline ingest data from several upstream processes and feed them to the relevant downstream systems. This means that the performance of several downstream processes depends on the stability of the relevant upstream processes and failures are hard to isolate without proper domain expertise. \par
\begin{figure}[!h]
\begin{center}
\includegraphics[ width=0.3\textwidth]{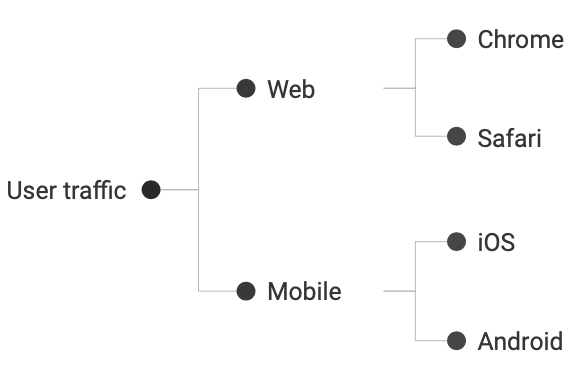}
\caption{User traffic tree by data types}
\label{flow}
\end{center}
\end{figure}

For instance, let us consider the user traffic tree shown in Figure \ref{flow}. This tree essentially shows how user traffic data can be segmented to corresponding devices and can be further segmented to several operating systems or browser types, and so on. Now, a failure at any of the node or layer of this tree will cause an impact that will propagate up to the root node of the tree. For example, if a bug is introduced in any android release that can cause a significant effect on user traffic, many downstream processes will follow varying impact if they directly or indirectly ingest data from either android user traffic or its parent/children pipelines. This tree structure can easily become very complex in a moderately large software system and getting any insights about the actual problem becomes almost impossible.\par
In this paper, we consider an automated root cause detection method that can operate over many time series collected from interdependent systems. We consider time series observed in several scales and map the residual process into a sequence of discretized states to have them in a comparable format. We consider a linear state space model to understand explainable patterns and HMM based state estimation for the residual process. Several other structural models like ARIMA can be used if the data shows a strong correlation and cyclical pattern. After identifying the state sequences, we perform a DTW based temporal alignment to group the similar residual state propagations into clusters. Hence if a set of anomalies is observed, it can be automatically grouped with similar patterns in terms of causality score \cite{li2015dynamic} which significantly reduces the size of the relevant feature dimensions to explore. Our technique focuses on the root cause detection of anomalous sub series and assume that the anomalies have already been detected. This is a key assumption that reduces a huge computational cost specifically for high-frequency data monitoring, where the technique only focuses on the problematic time segments and relies on the anomaly classification based on a standard anomaly detection method running somewhere upstream. 
\section{Background and Related Works}
When dealing with detecting anomalies in a real and practical application, it is not only important to isolate anomalous data points from the whole series of observations, but also to isolate them on an event level. For instance, a simple anomaly detection process can find a drop in the time series related to the volume of a website traffic. However, to get any meaningful insight about the observed anomaly, we should be able to move one step forward and attach that drop to a specific event (e.g. a possible bug in a new software version). To this end, several methods have been proposed to perform anomaly detection on a single time series to efficiently identify the problematic regions. \cite{malhotra2015long} explored a deep learning approach based on LSTM (Long Short-term Memory) for anomaly detection. \cite{zhang2003time} proposed a hybrid of ARIMA and neural network model that gains a better precision of forecasting by considering any non-linear relation over time. \par
While such methods can be beneficial for ad-hoc use cases, researchers have identified the need for considering associations between multiple time series in root cause detection of complex software pipelines. \cite{krishnamurthy2014scalable} proposed a Bayesian network based causality detection method for several cyber and physical features of the system. \cite{silveira2010urca} introduced an unsupervised method where anomalies are isolated by successively reducing the anomalous feature space. \cite{brauckhoff2009anomaly} used a histogram based approach where several underlying features are binned and a histogram based association is applied using the KL divergence technique. \par
Another way of looking into the root cause detection method is to find data driven alignment between the different time series under consideration. This technique is specifically useful when no information is available regarding the dependencies of the underlying processes generating the time series. \cite{kassidas1998synchronization}, \cite{ramaker2003dynamic} discussed a time series alignment approach using dynamic time warping (DTW) that stretches or shrinks a given time series with respect to another time series with very similar patterns and also generates a cumulative cost that represents the similarity strength between the two time series. This idea has been extended in \cite{li2015dynamic} where the authors have proposed a DTW based method to find causality of anomalies between a pair of nonstationary time series. They considered a Dynamic latent variable model and a reconstruction based contribution to obtain the faulty variables and then implemented clustering to incorporate DTW based similarity within a cluster. \cite{liu2016unsupervised} and \cite{liu2016root} has extended the idea of causality based root cause detection by addressing it's variation over the system. They proposed a root cause detection method using sequential state switching and artificial anomaly association based on Restricted Boltzman Machine. \cite{kim2013root} have proposed MonitorRank for root cause detection over large software systems by incorporating several time series metrics along with the call graphs between several services into the model.\par
A relatively less explored area of research is to analyze system wide anomalous patterns in related time series where the time series themselves may not be aligned but a causal pattern can be drawn from the temporally lagged fluctuations. A problem on an upstream process can cause anomalies in the several dependent processes which should cause a temporal misalignment if the time series is granular enough to distinguish between the action and the reaction phase. Moreover, a complex software system has several such parent and dependent processes on each level of the dependency graph and hence the time series under consideration may not be homogeneous throughout the whole system. Different groups of related processes (and the corresponding time series) can be in different ``states'' and hence, it is important to group them using the anomaly patterns and isolate any problems within the groups. \par
\section{Root Cause Detection Method}
\subsection{Time series as sequential states}
The challenges in root cause detection of observed anomalies over multiple time series come from several aspects:
\begin{enumerate}
\item The anomalies or the structural patterns of the time series are not temporally aligned, hence, it is not feasible to use a simple correlation metric.
\item Warping several time series is not trivial as different processes generate time series in different scales, making it difficult to define an ideal cost function.
\item The time series representing the underlying processes are not homogeneous in nature in order to isolate a specific irregular pattern.
\end{enumerate}
To address the second challenge, \cite{li2015dynamic} have proposed a DTW based causality approach where they have addressed the scaling problem by normalizing all the time series into homogeneous scale between $0$ to $1$. However, this solution over-penalizes the normal (i.e. non-anomalous) observations if an extreme anomaly is observed in the time segment. \par
Essentially, the structural similarity between a set of time series is a local phenomena rather than a global one. Our proposed method revolves around the idea that the observations of a given time series is an \textit{emitted} value of the underlying process at that specific time. In order to elaborate our idea, let us denote $Y_{i1}, Y_{i2}, . . ., Y_{iT}$ as the sequential observations from a given process $(S_{i})$ which has a set of internal and external dependencies denoted by $\mathcal{D}_{i} = \{d_{ij}, j\in {1,2,. . ., p}\}$ where $d_{ij}$ is the $j^{th}$ observed dimension for the $i^{th}$ process. \par
Let us consider $\{X_{ij}, j\in 1(1)T\}$ is the hidden state corresponding to the observed state $y_{ij}$. Hence, we can define a simple linear model as $Y_{it} = A_{it}X_{it} + V_{it}$. Here $A_{it}$ is the relationship matrix and $V_{it}\sim N(0, R)$ is the measurement noise. Further, we consider formulating the relationship between two consecutive states as $X_{it+1} = C_{it}(\mathcal{D}_{i})X_{it} + W_{it}(\mathcal{D}_{i})$. Here $C_{it}(\mathcal{D}_{i})$ defines the relationship structure between the consecutive states as well as the off-diagonal elements represents any cross-sectional relationships. $W_{it}(\mathcal{D}_{i})\sim N(0,Q)$ represents the process noise for the underlying state propagation. It is important to note that, the model assumes both the model coefficient and the error process being a function of the underlying dimensions.
\subsection{Discretized Residual Process}
The key assumption of the modeling technique is the status of the underlying residual process. With an assumption of optimal representation of $Y_{it}$ by $X_{it}$, we can rewrite the linear state space model defined in the last section as:
\begin{eqnarray}
\label{eq1}
Y_{it} = E_{it} + V_{it}
\end{eqnarray}
where,
\begin{eqnarray*}
E_{it} = A_{it}X_{it}
\end{eqnarray*}
Equation \ref{eq1} has two components. The first one, $E_{it}$, comes from an explainable linear form and the second source, $V_{it}$, is the unexplainable emission error process:
\begin{eqnarray}
\label{eq2}
V_{it} = \mathcal{E}_{it} + \Psi_{it}
\end{eqnarray}
Here $\mathcal{E}_{it}$ represents the uncontrollable emission error and $\Psi_{it}$ represents the unexplained errors related to a specific dimension combination $\mathcal{D}_{i} = D_{i}$. In the ideal situation, where all the processes are under control, we expect $||\Psi_{it}||_{2} < \delta$ for a small enough $\delta > 0$, otherwise, we expect a large enough $||\Psi_{it}||_{2}$ in case of an anomaly. \par
One major challenge in our method is to relate the $\Psi_{i}$ for different $i$'s to obtain a causal relationship for any fluctuations within the set of $\{S_{i}: \forall i\}$. To that end, we take a strong assumption that, error process $\Psi_{it}$ always propagates through some ordered discretized states while transitioning from one innovation to another (i.e. from $X_{t}$ to $X_{t+1}$). If we assume $\mathcal{E}_{it}$ to be a $0$ mean error process, we can write:
\begin{eqnarray}
\label{eq3}
E(V_{it}) = 0 + E(\Psi_{it}) = E(\Psi_{it})
\end{eqnarray}
Let us consider $\mathcal{G}: \Psi_{t} \longrightarrow \xi_{t}$ be the mapping from the (unexplained) error process to the discretized states. Here, $\xi_{t}\in \{\Omega_{k}: k\in 1(1)K\}$ is an ordered set of discretized states. The key assumption behind this formulation is the localization structural pattern of the time series that induces a uniform robust scaling. As described in the several difficulties in root cause detection in section 2, this formulation defines the discretized states on top of the local propagation of the error process and is not susceptible to any global phenomena that might influence the local fluctuations.  \par
\begin{figure}
\begin{center}
\includegraphics[ width=0.5\textwidth]{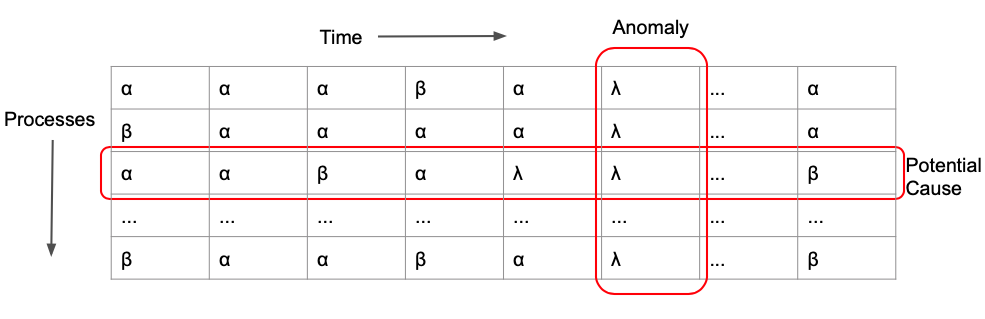}
\caption{Discretized ordered state sequence for the unexplained error processes. $\alpha$: normal state, $\beta$: warning state, $\lambda$: anomalous state}
\label{fig3}
\end{center}
\end{figure}
The main advantage to represent the unexplained error process as above is that it brings all the considered time series into a uniform comparable scale. For example, Figure \ref{fig3} shows the discretized states of the unexplained error processes for the corresponding set of time series which enters the anomalous state $\lambda$ at a certain time point and we can clearly narrow down to a specific process that temporally precedes the inheritance of the anomalous state compared to its peers. We can think of this situation as very similar to the standard Gaussian bound where values between $-3$ and $3$ are treated within normal bounds (or within the $3\sigma$ limits) and anything beyond that can be treated as outliers. A further more sophisticated formulation can be performed by defining a Hidden Markov Model where the residual process can be mapped to a set of ordered discretized states.
\subsection{Temporal Alignment}
In general, in case of a faulty variable or an event in a software system, the anomaly first appears in a small set of processes and then propagates to a larger related group. In other word, we expect a temporal lag in the anomaly occurrence if the time series is observed in a granular enough frequency. Also, processes with moderately extensive overlapping properties tend to show similar structural patterns and have a very conforming underlying state propagation as well. Hence, we consider implementing the dynamic time warping (DTW) technique which is a very effective method for such time series alignment. \par
Let us consider two processes $S_{U}$ and $S_{V}$ and suppose the corresponding discretized error states are $\{\xi_{Ut}, t=T_{1}(1)T_{2}\}$ and $\{\xi_{Vt}, t=T_{1}(1)T_{2}\}$. Suppose these states represents a time segment $(T_{1}, . . ., T_{2})$ containing an anomalous event. Now, DTW runs alignment in such a way that if we consider a cost matrix that defines the distance (e.g. Euclidean distance) between any two points in the time series, the method finds an optimal path in the matrix grid that minimizes the cumulative cost to get from one end of both of the time series to the other. In other words, the optimal path $\mathcal{F}$ is an optimal sequence of consecutive points in the matrix such that:
\begin{eqnarray*}
\mathcal{F} = \{w(1), w(2), . . ., w(L)\},\hspace{0.1cm}where\hspace{0.1cm}T_{2}-T_{1}\leq L\leq 2(T_{2}-T_{1})
\end{eqnarray*}
Here, $w(l)$ is a point in the matrix.
\begin{figure}
\begin{center}
\includegraphics[ width=0.45\textwidth]{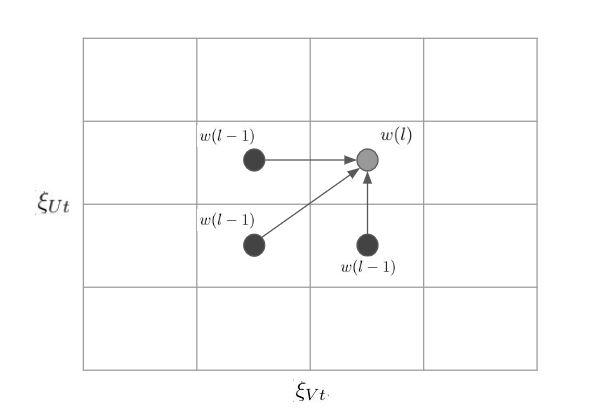}
\caption{Symmetric local constraint}
\label{fig4}
\end{center}
\end{figure}
We consider a symmetric local constraint while constructing the path $\mathcal{F}$ defined in Figure $\ref{fig4}$. That means, we allow any horizontal, vertical and diagonal movements while constructing the path. The end goal is to find the optimal path such that it minimizes the cumulative distance:
\begin{eqnarray*}
D_{opt}(\xi_{U}, \xi_{V}) = \min_{\mathcal{F}}\Sigma_{l=1}^{L}d[w(l)]
\end{eqnarray*}
where $d(\cdot)$ is some distance measure.
Moreover, as we like to see if the anomalous pattern has appeared in one time series before the other, we follow a similar approach to that described in \cite{li2015dynamic} where they have suggested a shift operator $\tau$ as:
\begin{eqnarray}
D_{opt}(\tau) = D_{opt}(\xi_{U}[1: T_{2}-T_{1}-\tau[, \xi_{V}[\tau + 1:T_{2}-T_{1}])
\end{eqnarray}
Hence, the causality index can be defined as:
\begin{eqnarray}
DCI(\xi_{U}, \xi_{V}) = \frac{\min_{w}D_{opt}(\tau)}{D_{opt}(0)}
\end{eqnarray}
Since we can compute $D_{opt}(\tau;U, V)$ for every possible pair of processes $S_{U}$ and $S_{V}$, we can essentially get a pairwise optimal distance and then can run a simple K-means like clustering algorithm using $D_{opt}(\tau;U, V)$ and every cluster will store the most similar time series with respect to their regular as well as the anomalous pattern. However, the anomalous time segment will put more weight towards the anomalous region while the alignment compared to the non-anomalous region. We can outline the clustering technique using DTW in the following algorithm:
\begin{algorithm}
\caption{SAC-DTW based clustering}
\label{kmeans}
\begin{algorithmic}[1]
\State Pre-compute $D_{opt}(\xi_{U}, \xi_{V}), DCI(\xi_{U}, \xi_{V})$ $\forall\hspace{0.1cm} U\hspace{0.1cm} and\hspace{0.1cm} V$
\While{$M \leq M_{max}$}
\While {$k \leq iter_{max}$}
\State \{$\xi_{ci}^{(k)}: i=1(1)M$\} $\gets$ Random cluster centers for clusters \{$\mathcal{C}_{i}^{(k)}: i=1(1)M$\}
\State $Assign\hspace{0.1cm} \xi_{j}\hspace{0.1cm} to\hspace{0.1cm} optimal\hspace{0.1cm} \mathcal{C}_{i}^{(k)}\hspace{0.1cm}\forall j \hspace{0.1cm}w.r.t \hspace{0.1cm}D_{opt}(\cdot , \cdot)$
\EndWhile
\State Obtain the optimal \{$\mathcal{C}_{i}^{(k)}: i=1(1)M$\} w.r.t $k$
\EndWhile
\State Find\hspace{0.1cm} M\hspace{0.1cm} that\hspace{0.1cm} minimizes\hspace{0.1cm} within\hspace{0.1cm} cluster\hspace{0.1cm} distances
\State $\min_{U,V\in \mathcal{C}_{i}}D_{opt}(\tau;U, V) \gets Causality\hspace{0.1cm} score\hspace{0.1cm} for\hspace{0.1cm} \mathcal{C}_{i}$
\end{algorithmic}
\end{algorithm}
A very small causality score (close to $0$) means that the subset of the dimension set represented through the processes contains the root cause of the problem. On the other hand, A high enough causality score (close to $1$) means the anomalies observed are the results of a propagation effect generated from a different cluster.
\section{State Estimation}
We use Kalman Smoothing as an estimation method for the State Space model. A filtering or structural state formulation could be used as well for specific applications. Here, the problem is to estimate the states given $Y_{1}, Y_{2}, . . ., Y_{T}$ (ignoring the subscript $i$ to consider any arbitrary process). The estimation technique is carried out in two steps for a specific Kalman Filter based method: prediction and correction. In the prediction step, we want to predict the next state of the process given the previous state:
\begin{eqnarray}
\hat{X}_{t+1|t} = C\hat{X}_{t|t}\\
P_{t+1|t} = CP_{t|t}C^{T} + Q
\end{eqnarray}
where $P_{t+1|t}$ and $P_{t|t}$ are the priori and the posteriori error covariances respectively. \par
After observing the actual value of $Y_{t+1}$, we take the correction step of our estimates as:
\begin{eqnarray}
K_{t+1} = P_{t+1|t}A^{T}(AP_{t+1|t}A^{T} + R)^{-1}\\
\hat{X}_{t+1|t+1} = \hat{X}_{t+1|t} + K_{t+1}(Y_{t+1} - A\hat{X}_{t+1|t})\\
P_{t+1|t+1} = P_{t+1|t} - K_{t+1}AP_{t+1|t}
\end{eqnarray}
Here $K_{t+1}$ is the Kalman gain at time $t+1$. \par
Now, to obtain the smoothed states, we need to estimate the state $X_{t|T}$ for $0\leq t\leq T$. Hence, we can simply outline the backward pass recursion as:
\begin{eqnarray}
\label{eq_smooth}
L_{t} = P_{t|t}C^{T}P_{t+1|t}^{-1}\\
\hat{X}_{t|T}=\hat{X}_{t|t} + L_{t}(\hat{X}_{t+1|T}-\hat{X}_{t+1|t})\\
P_{t|T} = P_{t|t} + L_{t}(P_{t+1|T}-P_{t+1|t})L_{t}^{T}
\end{eqnarray}
Equation $\ref{eq_smooth}$ essentially provides the smoothed states along with the error covariance structure. Referring back to equation \ref{eq1}, we can obtain an underlying residual process corresponding to the smoothed states as:
\begin{equation}
\Upsilon_{t} = Y_{t} - \hat{X}_{t|T}
\end{equation}
From equation \ref{eq3}, we can say $\Upsilon_{t}$ is the residual process corresponding to the smoothed states is an accurate representation of the actual unexplained error process $\Psi_{t}$. Simple thresholding based on the error covariance $P_{t|T}$ or HMM can be used for discretized state estimation in order to run DTW.
\begin{figure*}
\begin{center}
\subfloat{\includegraphics[ width=0.4\textwidth]{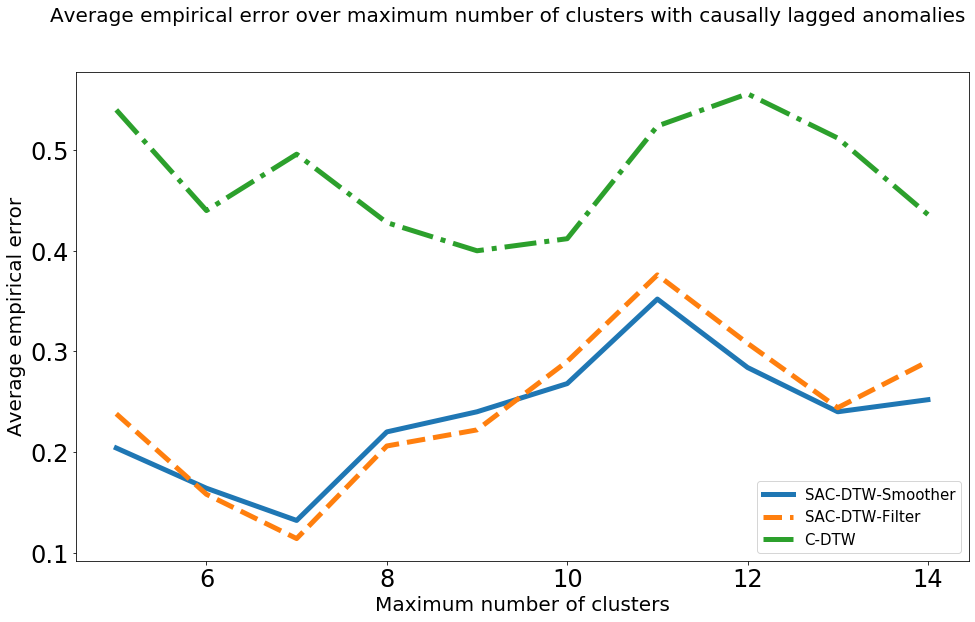}}
\subfloat{\includegraphics[ width=0.4\textwidth]{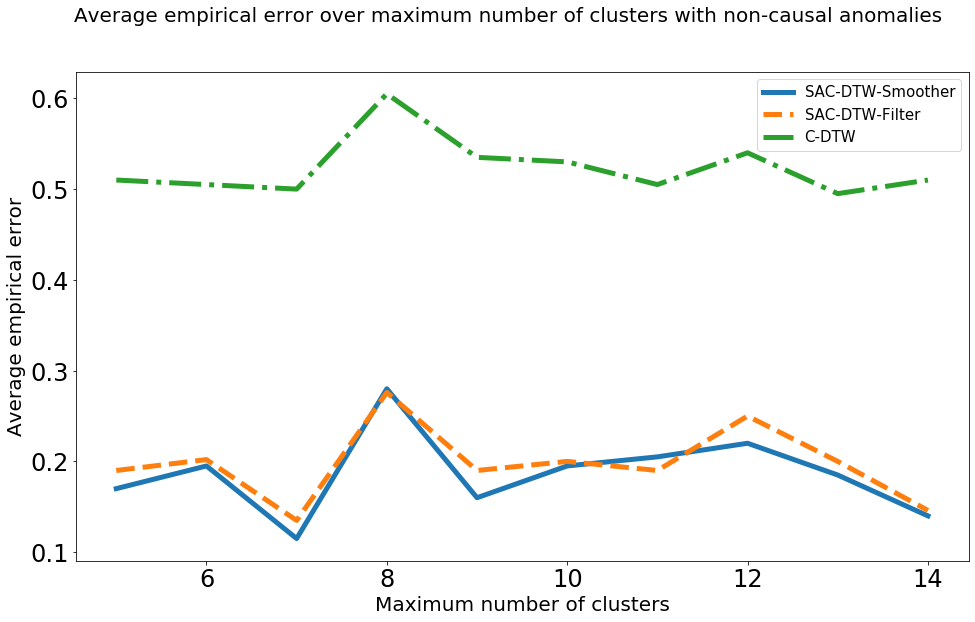}}
\caption{Average (over different anomaly intensities) empirical classification error with respect to the maximum number of clusters. The underlying time series are injected with artificial anomalies with causal lags (left figure) and no causality (right image).}
\label{cerror}
\end{center}
\end{figure*}
\begin{figure*}
\begin{center}
\subfloat{\includegraphics[ width=0.4\textwidth]{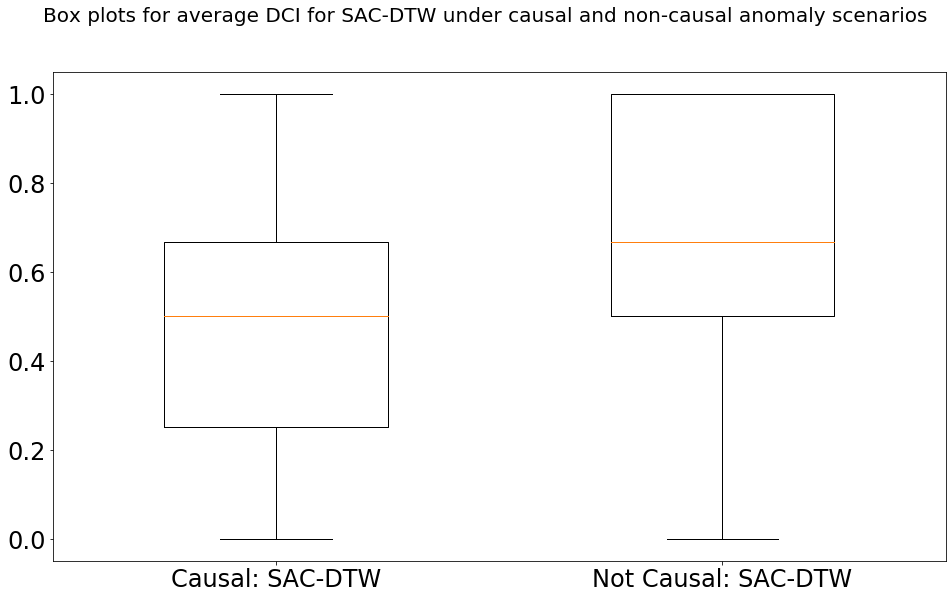}}
\subfloat{\includegraphics[ width=0.4\textwidth]{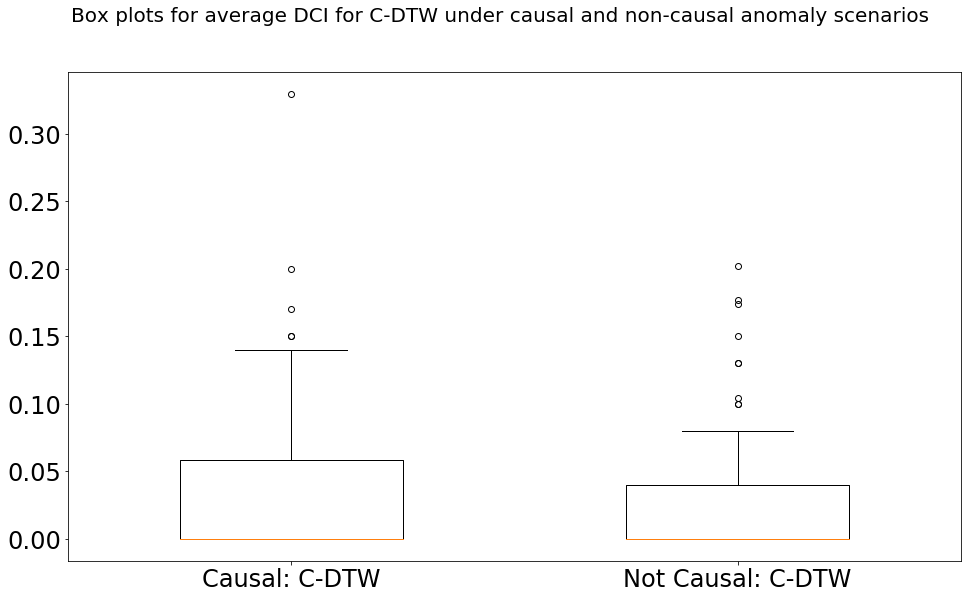}}
\caption{Box plot of causality index for SAC-DTW vs C-DTW under causal and non-causal scenarios}
\label{boxp}
\end{center}
\end{figure*}
\section{Simulation Study}
In order to perform empirical comparisons between the proposed temporal state alignment techniques in this paper (SAC-DTW with both the Kalman filter and smoother as the state estimation methods) and the existing DTW based causality detection method (C-DTW) (\cite{li2015dynamic}), we simulate two sets of $50$ time series from $5$ different groups each. Each group contains $10$ time series of length 50. We artificially generate anomalies in a specific temporal location unique for each group. In order to test the performance of causality detection of the proposed method, we allow temporal lag between the anomalies in each group from the first set. We allow the scaling factor to vary between considerable ranges (up to a factor of $10^{5}$) to incorporate the effect of states on local variations in different scales.  \par
Figure \ref{cerror} shows the performance of the SAC-DTW and C-DTW with respect to the classification error in terms of the cluster allocation. The test has been performed with respect to different numbers of cluster-max as the hyperparameter of the clustering technique. The classification error for SAC-DTW is much lower for different cluster max's compared to the C-DTW, although the classification error seems to increase very slowly in the case of causally lagged anomalies with the increase of cluster max. \par
Figure \ref{boxp} shows the performance of the SAC-DTW and C-DTW with respect to the causal and non-causal scenarios. There is a clear distinction between the causally related anomalies and the non causal anomalies with respect to the causality score for SAC-DTW. The distribution for the causally related anomalies is much closer to zero compared to the non-causal scenario, whereas the performance of C-DTW in both causal and non-causal cases are quite indistinguishable with very low scores of causality. \par
This section clearly shows the improvement in the performance when the underlying states are considered for isolating the causalities instead of the scaled raw scores. The clustering technique clearly helps to divide the problems into several subgroups and each of the subgroups can be further analyzed using their causality scores to isolate the most similar time series which carries temporal causality in the anomaly propagation.
\section{Applications}
Monitoring data quality plays a significant role in maintaining software systems and data pipelines within Zillow Group. To achieve this goal, we have designed and implemented an in-house data quality platform called \textit{Luminaire}. Our platform can collect data quality metrics over a period of time and train the best model that matches the characteristics of the data. We then score the observed metrics according to the trained model and send notification to the proper stakeholders if an anomaly is observed. Several critical operations within Zillow Group rely on \textit{Luminaire} to monitor the health of their pipelines as well as data quality metrics. Such operations include home value estimation (\textit{Zestimate}), home recommendation, real estate transactions, \textit{Premier Agent} and \textit{Zillow Offer} products. Due to the critical role of \textit{Luminaire} in the data ecosystem of Zillow Group, it is quite important to provide actionable anomaly alerts by isolating the root cause of a problem, where possible. This greatly reduces human involvement and brings more automation to maintaining system stability. The biggest challenge towards this goal however, is to generate a deterministic dependency mapping between the different process which is almost impossible to obtain due to its complexities and ever-evolving nature. Hence, it is important to derive a data driven understanding of the fault propagation in case of a failure.\par
For our empirical analysis, we specifically focus on monitoring clickstream data, which is an important part for Zillow's business operations as different downstream processes ingest this data in either raw or aggregated form. It is quite difficult to track down the root cause of the problem in case of an irregular fluctuation within clickstream or somewhere downstream due to the complex nature of clickstream data pipelines. Therefore, the idea of the root cause detection technique is to process a data driven signal and to isolate the root cause of the problem into the smallest possible subset of dimensions. \par
We take two key assumptions for a proper implementation of the proposed root cause detection method described in the paper:

\begin{enumerate}
\item The data is collected and aggregated in a granular enough scale in terms of time such that the temporal lag in the causality is empirically identifiable.
\item The data dimensions are granular enough so that the causal effect between the dimensions are empirically identifiable. \par
\end{enumerate}
\begin{table}
\caption{Root Cause Detection for Clickstream data}
\label{data_an}
\noindent\begin{center}
\scalebox{0.8}{\begin{tabular}{|p{3cm}|p{1.2cm}|p{1.2cm}|p{1.2cm}|p{1.2cm}|}
\cline{2-5}
\multicolumn{1}{c|}{}& \multicolumn{2}{c|}{SAC-DTW} & \multicolumn{2}{c|}{C-DTW} \\
\hline
Device-experiance & DCI-Avg & Gini impurity & DCI-Avg & Gini impurity \\
\hline
Desktop Web (C)       &   \textbf{0.621} &\textbf{0.368} &  0.961 & 0.443 \\
\hline
Real Estate - Android Phone Instant (NC)    &  \textbf{0.759} &\textbf{0.01} &  0.106 & 0.688\\
\hline
Real Estate - iPhone (C)   & 0.433 & \textbf{0.454} & \textbf{0.007} & 0.623 \\
\hline
Mortgage Marketplace - iPhone (NC)   & \textbf{0.855} & -  & 0.106 & - \\
\hline
Rentals - Android Tablet (C)   & \textbf{0.567} & \textbf{0.478} &  0.901 & 0.590 \\
\hline
Mortgage Marketplace - Android Phone (NC)    & \textbf{0.855} & \textbf{0.502}  & 0.007 & - \\
\hline
\end{tabular}}
\end{center}
\end{table}
In order to perform the empirical analysis, we consider clickstream data collected over a given time window where anomalies are present. The data under consideration contains several website related metrics (Visits, Home details page views, Authorized sign-ins etc.) and are collected for 6 different user device experience combinations specified in Table \ref{data_an}. The groups containing causal fluctuations denoted by (C) and non causal or no fluctuations denoted with $(NC)$ in the table. \par
Table \ref{data_an} shows the performance of the State aligned Causal DTW based root cause detection technique. First of all, the data contains 6 foundational clusters through the source, although, the observed anomalous patterns in the data further divides the data into several sub-clusters. Even though the computations have been done using the same max cluster ($=8$) for SAC-DTW and C-DTW, the former was able to find $5$ cluster which is closer to the truth than $4$ clusters found by the C-DTW technique. A further verification of the previous statement can be found from the column showing the Gini's impurity index for SAC-DTW which is moderately small for the groups with causal (C) fluctuations which proves the consistency of the clustering method compared to the C-DTW method. Moreover, SAC-DTW performed better in terms of finding the optimal number of clusters compared to the C-DTW method. We also observe DCI average for the SAC-DTW to be a strong indicator of the causality whenever the causality is present in the corresponding anomalous group, whereas the performance of C-DTW is quite random with respect to the causal and the non causal scenarios.
\section{Conclusions}
This paper introduced a method for performing root cause detection over temporally lagged fluctuations over several interlinked time series through the alignment of the underlying discretized hidden states of the residual process. This method performs clustering by running DTW to warp the discretized hidden states for the corresponding time series and assigns a causality score to the cluster that indicates a causal temporal lag between the fluctuations within the cluster. The experimental evaluation also shows the efficiency of the method in terms of clustering any causal patterns within the same group and isolating a causal group of time series using the causality index. This method has been shown to perform much robustly in terms of handling time series and fluctuations in different scales. In addition, while DTW is a computationally intensive process, this method is easy to scale through parallelization as the pairwise alignment can be applied independently (e.g. using distributed cross join operation in Apache Spark). 
\bibliographystyle{IEEEtran}
\bibliography{manuscript.bib}
\end{document}